\pdfoutput=1

\documentclass[11pt]{article}

\usepackage[]{emnlp2021}

\usepackage{times}
\usepackage{latexsym}
\usepackage{multirow}
\usepackage{float}
\usepackage{booktabs}
\usepackage{graphicx}
\usepackage{pgfplots}
\usepackage{enumitem}

\pgfplotsset{
tick label style={font=\small},
label style={font=\small},
title style={font=\normalsize},
legend style={font=\footnotesize}
}

\usepackage[T1]{fontenc}

\usepackage[utf8]{inputenc}

\usepackage{microtype}

\usepackage{cleveref}
\crefname{section}{\S}{\S\S}
\Crefname{section}{\S}{\S\S}
\crefname{table}{Table}{Tables}
\crefname{figure}{Figure}{Figures}
\crefname{algorithm}{Algorithm}{}
\crefname{equation}{eq.}{}
\crefname{appendix}{App.}{}
\crefformat{section}{\S#2#1#3}

\usepackage{float}

\usepackage{todonotes}
\makeatletter
\newcommand*\iftodonotes{\if@todonotes@disabled\expandafter\@secondoftwo\else\expandafter\@firstoftwo\fi} 
\makeatother

\newcommand{\xx}{\mathbf{x}}
\newcommand{\nn}{\mathbf{n}}

\newcommand{\nfew}{$N^{\textrm{\footnotesize{few}}}$}
\newcommand{\nmany}{$N^{\textrm{\footnotesize{many}}}$}
\newcommand{\methodname}{GE3}
\newcommand{\naug}{$n_\textrm{\footnotesize{aug}}$}

\title{Good-Enough Example Extrapolation}

\author{Jason Wei \\
  Google Research \\
  \texttt{jasonwei@google.com} \\}

\date{}

\def\shortpm{\hspace{0.4mm}$\pm$\hspace{0.4mm}}

\pgfplotsset{compat=1.17}
\begin{document}
\maketitle
\begin{abstract}
This paper asks whether extrapolating the hidden space distribution of text examples from one class onto another is a valid inductive bias for data augmentation. 
To operationalize this question, I propose a simple data augmentation protocol called ``good-enough example extrapolation'' (\methodname).
\methodname{} is lightweight and has no hyperparameters. 
Applied to three text classification datasets for various data imbalance scenarios, \methodname{} improves performance more than upsampling and other hidden-space data augmentation methods.


\end{abstract}

\section{Introduction}
Text classification is a fundamental task in NLP for which modern architectures have achieved high performance when training data is sufficient \cite{glue}.
In many applied settings where data collection and annotation is limited, however, a common challenge is data imbalance \cite{krawczyk2016learning}, in which training data from certain categories is scarce. 
A classic example of such a scenario is intent classification, where developers may wish to update a conversational agent to be able to classify new intents, but the amount of training data for these new intents lags behind that of existing ones \cite{Bapna2017,gaddy2020overcoming}. 

One common method for mitigating the weaknesses of limited training data is data augmentation, a paradigm that is become increasingly seductive in the NLP landscape \cite[see][for a survey]{feng2021survey}. 
While data augmentation may be of only incremental utility when training data is sufficient \cite{longpre-etal-2020-effective}, it can be particularly helpful for mitigating data scarcity in low-resource settings such as few-shot classification \cite{wei-fewshot} or, as this paper will soon explore, data-imbalanced text classification.
\begin{figure}[ht]
    \centering
    \includegraphics[width=\linewidth]{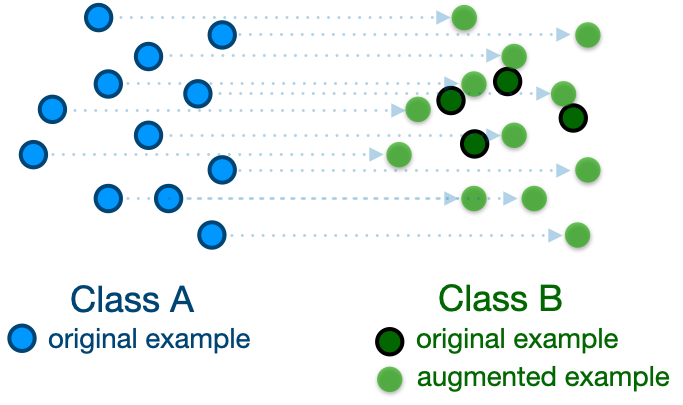}
    \vspace{-7mm}
    \caption{Diagram illustrating how \methodname\ extrapolates the hidden-space distribution of examples in class A onto class B.}
    \label{fig:pull}
    \vspace{-3mm}
\end{figure}


In this paper, I propose a simple data augmentation protocol called \textit{good-enough\footnote{Paying good respects to ``good-enough compositional augmentation'' \citep{andreas-2020-good}.} example extrapolation} (\methodname{}) for the class-imbalanced scenario. 
As shown in \cref{fig:pull}, \methodname\ extrapolates the hidden-space distribution of examples from one class onto another class.
\methodname\ has no hyperparameters, is model-agnostic, and requires little computational overhead, making it easy-to-use. 
In empirical experiments, I apply \methodname\ to intent classification, newspaper headline classification, and relation classification, finding that in a variety of class-imbalanced scenarios, \methodname\ substantially outperforms upsampling and other hidden state augmentation techniques.

\section{Hidden Space Extrapolation}
\paragraph{Intuition.}
Representation learning aims to map inputs into a hidden space such that desired properties of inputs are easily extractable from their continuous representations \cite{pennington-etal-2014-glove}. 
For many representation learning functions, inputs with similar properties map to nearby points in hidden space, and the distances between hidden space representations represent meaningful relationships \cite{mikolov-etal-2013-linguistic,10.1162/tacl_a_00327}.

\begin{figure}[h]
    \centering
    \includegraphics[width=\linewidth]{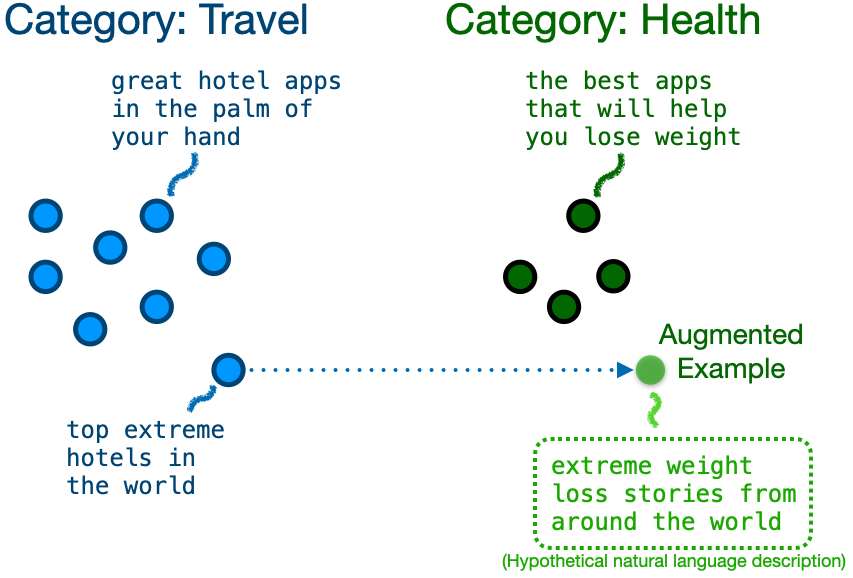}
    \caption{Intuition of extrapolating examples in hidden space from a travel category to a health category. A hypothetical natural language representation is shown for the augmented example generated via \methodname.}
    \label{fig:example}
\end{figure}

So for a given classification task, inputs from the same category will have some distribution (i.e., cluster) in hidden space, where the distance between distributions represents the relationship between categories, and the distribution of points within the same category models some random variable. 
\methodname\ leverages this intuition by extrapolating the distribution of points in the same category, which models some random variable, from one category onto another.
\cref{fig:example} illustrates this intuition in a hypothetical instance from the HuffPost dataset of news headlines classification, where the hidden-space relationship between two examples in the travel category can be extrapolated to form a new example in the health category.

\paragraph{Method.}
Formally, I describe \methodname\ as follows.
Given a text classification task with $k$ output classes $\{c_j\}_{j=1}^{k}$, denote the $n_c$ training examples in class $c$ as $\{X_c^i\}_{i=1}^{n_c}$ with corresponding hidden space representations $\{\xx_c^i\}_{i=1}^{n_c}$.
For each class $c$, let $\mu(c) = \frac{1}{n_c} \sum_{i=1}^{n_c}\xx_c^i$ indicate the mean of all hidden space representations in that class.

\methodname\ generates augmented examples by extrapolating the data distribution from a source class $c_{s}$ to a target class $c_{t}$ in hidden space.
And so for each hidden space representation $\xx_{c_s}^i$ in the source class, I generate a corresponding augmented example $\hat{\xx}_{c_t}$ in the target class:
\begin{equation}
    \hat{\xx}_{c_t} = \xx_{c_s}^i - \mu(c_s) + \mu(c_t) \ .
\end{equation}
In total, for each class in the training set, I can generate a set of extrapolated points from every other class, augmenting the size of the original training set by a factor of $k$.
I then train the classification model on the union of original data and extrapolated examples.
Notably, this extrapolation method operates without any hyperparameters, as augmented examples are generated via distributions from other classes instead of a noising function (c.f. augmentation techniques that usually have a strength parameter \cite{,sennrich-etal-2016-improving,wei-zou-2019-eda}).



\section{Experimental Setup}

I evaluate the proposed hidden space extrapolation protocol in several data imbalance scenarios on three diverse text classification datasets.

\subsection{Datasets}
\paragraph{SNIPS.} The Snips Voice Platform dataset\footnote{I used the processed data from \url{https://github.com/MiuLab/SlotGated-SLU/tree/master/data/snips}.} \cite{coucke2018snips} is an intent classification dataset that maps utterances to 7 different intents (e.g., `play music', `get weather', etc.). Each intent has about 1800 training examples. 
\vspace{-0.2em} \paragraph{HUFF.} The HuffPost dataset \cite{huff} comprises news headlines published on HuffPost from 2012--2018. Headlines are categorized into one of 41 classes (e.g., `health', `travel', etc.), and I split the dataset such that the training set has 700 examples per class. 
\vspace{-0.2em} \paragraph{FEWREL.} The few-shot relation classification dataset \cite{han-etal-2018-fewrel} contains categorized relationships between specified tokens (e.g., ‘capital of,’ ‘birth name,’ etc). 
The posted training set contains 64 classes, and I perform a train-test split such that each class has 500 examples in the training set and 100 examples in the evaluation set.

For all three datasets, I create artificially imbalanced datasets via random sampling. Specifically, I randomly select half the classes to maintain the original number of examples \nmany\ (i.e., \nmany\ $ = \{1800, 700, 500\}$ for SNIPS, HUFF, and FEWREL respectively), and for the other half of the classes, I train on only a subset of \nfew\ examples. I run experiments on a range of \nfew.

\begingroup
\newcommand{\sentry}[2]{#1\scriptsize{\shortpm#2}}
\newcommand{\sentryb}[2]{\textbf{#1}\scriptsize{\shortpm#2}}
\begin{table*}[ht]
    \centering
    \small
    \begin{tabular}{l | c c | c c | c c | c c }
        \toprule
        & \multicolumn{2}{c|}{SNIPS ($k=7$)} & \multicolumn{2}{|c|}{HUFF ($k=41$)} & \multicolumn{2}{|c|}{FEWREL ($k=64$)}  \\
        & $N^\textrm{\scriptsize{few}}$=20 &  $N^\textrm{\scriptsize{few}}$=50 & $N^\textrm{\scriptsize{few}}$=20 &  $N^\textrm{\scriptsize{few}}$=50 & $N^\textrm{\scriptsize{few}}$=20 &  $N^\textrm{\scriptsize{few}}$=50 & \underline{Average} & $\Delta$ \\
        \midrule
        Baseline (upsampling) & \sentry{88.9}{1.1} &  \sentry{92.1}{0.8} & \sentry{27.3}{0.2} & \sentry{30.3}{0.3} & \sentry{49.6}{0.6} & \sentry{56.7}{0.6} & 55.3 & - \\
        \midrule
        Interpolate & \sentry{89.3}{1.1} & \sentry{92.0}{0.9}  & \sentry{27.0}{0.1}  & \sentry{29.9}{0.2}  & \sentry{49.1}{0.5}  & \sentry{56.1}{0.5} & 55.1 & -0.2 \\
        Within-extrapolation & \sentry{86.5}{1.2} & \sentry{90.5}{0.8}  & \sentry{28.7}{0.2}  & \sentry{30.8}{0.3}  & \sentry{49.2}{0.3}  & \sentry{56.0}{0.6} & 54.8 & -0.5 \\
        Linear Delta & \sentry{86.4}{1.0} & \sentry{90.7}{1.1}  & \sentry{29.9}{0.2} & \sentry{32.4}{0.3} & \sentry{50.9}{0.6} & \sentry{58.2}{0.6} & 55.7 & +0.4 \\
        Uniform Noise & \sentry{88.8}{0.9} & \sentry{91.9}{0.8}  & \sentry{30.2}{0.8}  & \sentry{33.3}{0.2}  & \sentry{50.9}{0.6}  & \sentry{57.8}{0.4} & 56.6 & +1.3 \\
        Gaussian Noise & \sentry{89.1}{1.1} & \sentry{92.1}{0.9}  & \sentry{31.9}{0.3}  & \sentry{33.7}{0.2}  & \sentry{53.0}{0.6}  & \sentry{60.3}{0.6} & 58.0 & +2.7 \\
        \textbf{\methodname\ (ours)} & \sentryb{90.6}{0.6} & \sentryb{92.8}{0.7} & \sentryb{32.7}{0.2} & \sentryb{36.8}{0.1} & \sentryb{56.3}{0.6} & \sentryb{64.0}{0.2} & \textbf{59.9} & \textbf{+4.6} \\
        \bottomrule
    \end{tabular}
    \caption{
    Accuracy (\%) of \methodname, upsampling, and five other hidden space augmentation techniques on data-imbalanced text classification scenarios, where half of the classes are restricted to \nfew\ training examples. $k$: number of total classes for a classification task. $\Delta$: improvement over the upsampling baseline.
    }
    \label{tab:main_table}
\end{table*}
\endgroup
\subsection{Model and Experimental Procedures}
For the classification model, I use BERT encodings \cite{devlin-etal-2019-bert} with max-pooling  \cite{reimers-gurevych-2019-sentence} to generate sentence embeddings and add an additional softmax layer for classification. 
I implement \methodname\ at this final max-pooled hidden layer, which has size 768.
That is, the hidden-space augmentation method only updates classifier weights after the BERT encoder.
Before training, the data processing pipeline upsamples from classes with fewer examples until all classes have the same number of examples in the training set.
I run all experiments for five random seeds.

\subsection{Hidden Space Augmentation Baselines}
As baselines for comparison, I also explore several other hidden space augmentation techniques:
\paragraph{Example interpolation.} 
Given the hidden space representations of two examples $\xx^i_c$ and $\xx^j_c$\ in the same class, I generate an augmented example 
\begin{equation}
    \hat{\xx}_c = \frac{1}{2} (\xx^i_c + \xx^j_c)\ .
\end{equation}

\vspace{-0.3em} \paragraph{Within-extrapolation.} \cite{devries2017dataset,kumar-etal-2019-closer}. 
Given two examples $\xx^i_c$ and $\xx^j_c$ in the same class, I extrapolate the hidden space between the two to form an augmented example 
\begin{equation}
    \hat{\xx}_c = \lambda \cdot (\xx^i_c + \xx^j_c) - \xx^i_c\ .
\end{equation}
Following \citet{kumar-etal-2019-closer}, I use $\lambda=0.5$.

\vspace{-0.3em} \paragraph{Linear delta.} \cite{kumar-etal-2019-closer}. 
The difference between two examples $\xx^i_c$ and $\xx^j_c$ in the same class can be added to a third example $\xx^j_c$ to form an augmented example:
\begin{equation}
    \hat{\xx}_c = (\xx^i_c - \xx^j_c) + \xx^k_c \ .
\end{equation}

\vspace{-0.3em} \paragraph{Noising.} 
Given some example $\xx_c^i$, I add noise $\nn$ to yield an augmented example
\begin{equation}
    \hat{\xx}_c = \xx^i_c + \nn(\cdot) \ .
\end{equation}
For $\nn(\cdot)$, I explore both \textbf{Uniform Noise}, where each element is uniformly sampled from $[a, b]$, where $a=-0.1$ and $b=-0.1$, as well as \textbf{Gaussian Noise}, where each element is sampled from  $\mathcal{N}(\mu, \sigma)$, where $\mu=0$ and $\sigma=0.1$. 

\vspace{0.5em} 

\noindent For these techniques, I generate augmented examples until each class has \naug\ $\cdot$ \nmany\ training examples, where \naug\ $=5$ (a choice which is later explored in \cref{fig:n_aug}).
 
\section{Results}
\pgfplotsset{width=4cm,height=3.4cm,compat=1.9}
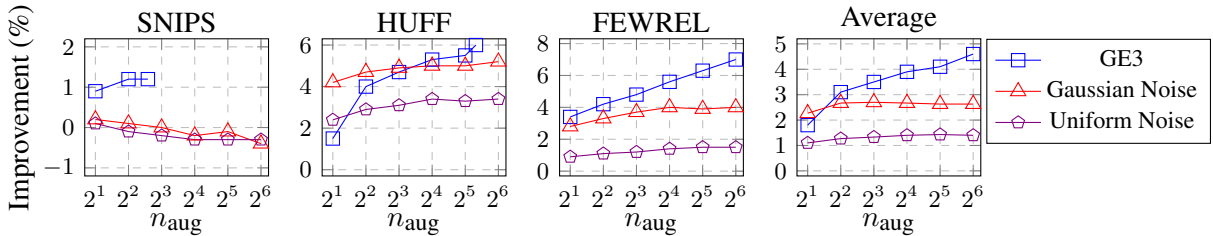
\begin{figure*}[th]
\begin{centering}
\begin{tikzpicture}
\begin{axis}[
    title={SNIPS},
    title style={yshift=-1.5ex,},
    xmode=log,
    log basis x={2},
    xlabel={$n_\textrm{\footnotesize{aug}}$},
    ylabel={Improvement (\%)},
    xmin=1.6, xmax=75,
    ymin=-1.2, ymax=2.2,
    xtick={2, 4, 8, 16, 32, 64},
    ytick={-1, 0, 1, 2},
    legend pos=south east,
    ymajorgrids=true,
    xmajorgrids=true,
    grid style=dashed,
    x label style={at={(axis description cs:0.5,-0.2)},anchor=north},
    y label style={at={(axis description cs:-0.215,0.5)},anchor=south},
]
\addplot[
    color=blue,
    mark=square,
    mark size=2.5pt,
    ]
    coordinates {
    (2,     0.9)
    (4,     1.2)
    (6,     1.2)
    };
\addplot[
    color=red,
    mark=triangle,
    mark size=3.3pt,
    ]
    coordinates {
    (2,     0.2)
    (4,     0.1)
    (8,     0.0)
    (16,    -0.2)
    (32,    -0.1)
    (64,    -0.4)
    };
\addplot[
    color=violet,
    mark=pentagon,
    mark size=2.5pt,
    ]
    coordinates {
    (2,     0.1)
    (4,     -0.1)
    (8,     -0.2)
    (16,    -0.3)
    (32,    -0.3)
    (64,    -0.3)
    };
\end{axis}
\end{tikzpicture}
\begin{tikzpicture}
\begin{axis}[
    title={HUFF},
    title style={yshift=-1.5ex,},
    xmode=log,
    log basis x={2},
    xlabel={$n_\textrm{\footnotesize{aug}}$},
    ylabel={},
    xmin=1.6, xmax=75,
    ymin=-0.3, ymax=6.3,
    xtick={2, 4, 8, 16, 32, 64},
    ytick={0, 2, 4, 6},
    legend pos=south east,
    ymajorgrids=true,
    xmajorgrids=true,
    grid style=dashed,
    x label style={at={(axis description cs:0.5,-0.2)},anchor=north},
    y label style={at={(axis description cs:-0.145,0.5)},anchor=south},
]
\addplot[
    color=blue,
    mark=square,
    mark size=2.5pt,
    ]
    coordinates {
    (2,     1.5)
    (4,     4)
    (8,     4.7)
    (16,    5.3)
    (32,    5.5)
    (40,    6)
    };
\addplot[
    color=red,
    mark=triangle,
    mark size=3.3pt,
    ]
    coordinates {
    (2,     4.2)
    (4,     4.7)
    (8,     4.9)
    (16,    5.0)
    (32,    5.0)
    (64,    5.2)
    };
\addplot[
    color=violet,
    mark=pentagon,
    mark size=2.5pt,
    ]
    coordinates {
    (2,     2.4)
    (4,     2.9)
    (8,     3.1)
    (16,    3.4)
    (32,    3.3)
    (64,    3.4)
    };
\end{axis}
\end{tikzpicture}
\begin{tikzpicture}
\begin{axis}[
    title={FEWREL},
    title style={yshift=-1.5ex,},
    xmode=log,
    log basis x={2},
    xlabel={$n_\textrm{\footnotesize{aug}}$},
    ylabel={},
    xmin=1.6, xmax=75,
    ymin=-0.3, ymax=8.3,
    xtick={2, 4, 8, 16, 32, 64},
    ytick={0, 2, 4, 6, 8},
    legend pos=south east,
    ymajorgrids=true,
    xmajorgrids=true,
    grid style=dashed,
    x label style={at={(axis description cs:0.5,-0.2)},anchor=north},
    y label style={at={(axis description cs:-0.145,0.5)},anchor=south},
]
\addplot[
    color=blue,
    mark=square,
    mark size=2.5pt,
    ]
    coordinates {
    (2,     3.4)
    (4,     4.2)
    (8,     4.8)
    (16,    5.6)
    (32,    6.3)
    (64,    7.0)
    };
\addplot[
    color=red,
    mark=triangle,
    mark size=3.3pt,
    ]
    coordinates {
    (2,     2.8)
    (4,     3.3)
    (8,     3.7)
    (16,    4)
    (32,    3.9)
    (64,    4)
    };
\addplot[
    color=violet,
    mark=pentagon,
    mark size=2.5pt,
    ]
    coordinates {
    (2,     0.9)
    (4,     1.1)
    (8,     1.2)
    (16,    1.4)
    (32,    1.5)
    (64,    1.5)
    };
\end{axis}
\end{tikzpicture}
\begin{tikzpicture}
\begin{axis}[
    title={Average},
    title style={yshift=-1.5ex,},
    xmode=log,
    log basis x={2},
    xlabel={$n_\textrm{\footnotesize{aug}}$},
    ylabel={},
    xmin=1.6, xmax=75,
    ymin=-0.2, ymax=5.2,
    xtick={2, 4, 8, 16, 32, 64},
    ytick={0, 1, 2, 3, 4, 5},
    legend pos=outer north east,
    ymajorgrids=true,
    xmajorgrids=true,
    grid style=dashed,
    x label style={at={(axis description cs:0.5,-0.2)},anchor=north},
    y label style={at={(axis description cs:-0.145,0.5)},anchor=south},
]
\addplot[
    color=blue,
    mark=square,
    mark size=2.5pt,
    ]
    coordinates {
    (2,     1.8)
    (4,     3.1)
    (8,     3.5)
    (16,    3.9)
    (32,    4.1)
    (64,    4.6)
    };
    \addlegendentry{\methodname}
\addplot[
    color=red,
    mark=triangle,
    mark size=3.3pt,
    ]
    coordinates {
    (2,     2.27)
    (4,     2.67)
    (8,     2.7)
    (16,    2.67)
    (32,    2.63)
    (64,    2.63)
    };
    \addlegendentry{Gaussian Noise}
\addplot[
    color=violet,
    mark=pentagon,
    mark size=2.5pt,
    ]
    coordinates {
    (2,     1.1)
    (4,     1.27)
    (8,     1.33)
    (16,    1.4)
    (32,    1.43)
    (64,    1.4)
    };
    \addlegendentry{Uniform Noise}
\end{axis}
\end{tikzpicture}
\caption{
Improvements over the upsampling baseline from augmentation methods based on how many duplicates of training data were made (\naug). 
\methodname\ extrapolates examples from one class to another, so it generates at most \naug$=6$ for SNIPS, \naug$=40$ for HUFF and \naug$=63$ for FEWREL. 
}

\label{fig:n_aug}
\end{centering}
\end{figure*}

\cref{tab:main_table} shows results for \methodname\ on the three datasets for \nfew\ $=20$ and \nfew\ $=50$.
\methodname\ outperforms the upsampling baseline by an average of 4.6\%, with strongest improvements on HUFF and FEWREL. 
Of the other augmentation techniques, Gaussian noising and uniform noising had the best performance, with an average improvement of 2.7\% and 1.3\%, respectively. 
Whereas these techniques only enforce smoothness around the distribution of points in a single class, I hypothesize that \methodname\ improved performance more because it injects a stronger inductive bias that the distribution of examples of the same class around their mean can be extrapolated to other classes.
\pgfplotsset{width=6.3cm,height=4.9cm,compat=1.9}
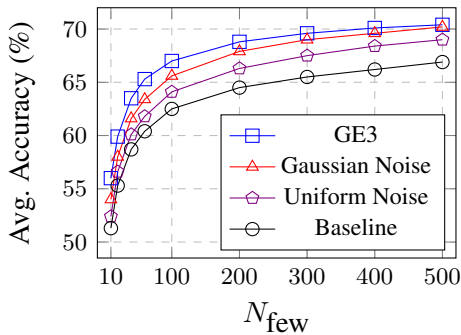
\begin{figure}[th]
\begin{centering}
\begin{tikzpicture}
\begin{axis}[
    title={},
    title style={yshift=-1.5ex,},
    xlabel={$N_\textrm{few}$},
    ylabel={Avg.\ Accuracy (\%)},
    xmin=-10, xmax=520,
    ymin=48.5, ymax=72,
    xtick={10, 100, 200, 300, 400, 500},
    ytick={50, 55, 60, 65, 70},
    legend pos=south east,
    ymajorgrids=true,
    xmajorgrids=true,
    grid style=dashed,
    x label style={at={(axis description cs:0.5,-0.135)},anchor=north},
    y label style={at={(axis description cs:-0.145,0.5)},anchor=south},
]
\addplot[
    color=blue,
    mark=square,
    mark size=2.5pt,
    ]
    coordinates {
    (10,    56.0)
    (20,    59.9)
    (40,    63.5)
    (60,    65.3)
    (100,   67.0)
    (200,   68.8)
    (300,   69.6)
    (400,   70.1)
    (500,   70.4)
    };
    \addlegendentry{\methodname}
\addplot[
    color=red,
    mark=triangle,
    mark size=2.5pt,
    ]
    coordinates {
    (10,    54.0)
    (20,    58.0)
    (40,    61.6)
    (60,    63.4)
    (100,   65.6)
    (200,   67.9)
    (300,   69.0)
    (400,   69.6)
    (500,   70.2)
    };
    \addlegendentry{Gaussian Noise}
\addplot[
    color=violet,
    mark=pentagon,
    mark size=2.5pt,
    ]
    coordinates {
    (10,    52.4)
    (20,    56.6)
    (40,    60.1)
    (60,    61.8)
    (100,   64.1)
    (200,   66.3)
    (300,   67.5)
    (400,   68.4)
    (500,   69.0)
    };
    \addlegendentry{Uniform Noise}
\addplot[
    color=black,
    mark=o,
    mark size=2.5pt,
    ]
    coordinates {
    (10,    51.3)
    (20,    55.3)
    (40,    58.7)
    (60,    60.4)
    (100,   62.5)
    (200,   64.5)
    (300,   65.5)
    (400,   66.2)
    (500,   66.9)
    };
    \addlegendentry{Baseline}
\end{axis}
\end{tikzpicture}
\vspace{-2mm}
\caption{
Comparing \methodname\ with the upsampling baseline, gaussian noise, and uniform noise for a range of \nfew\ (the number of training examples in the imbalanced classes). The performance shown is averaged over three datasets. 
}
\vspace{-3mm}
\label{fig:nfew}
\end{centering}
\end{figure}

Moreover, as \cref{tab:main_table} only shows results for \nfew\ $\in$ $\{20, 50\}$, in \cref{fig:nfew} I compare \methodname\ with upsampling, as well as Gaussian and uniform noise (the strongest baselines), for \nfew $\in$ $\{10$, $20$, $40$, $60$, $100$, $200$, $300$, $400$, $500\}$. 
\methodname\ improves performance across a wide variety of \nfew\ values, with improvements over the baselines slightly diminishing when training data is more balanced (as expected).

Finally, one of the reasons that \methodname\ improves performance more than other techniques could be that each class gets extrapolated examples from every other class, and if classes have unique distributions, then these extrapolated examples are valuable as additional training data.
Therefore, I perform an ablation study using a variable \naug, which restricts the number of other classes a given class can receive extrapolated examples from.
For instance, if \naug\ $=2$, then any given class may only receive extrapolated examples from two other random classes, even if there are 63 other classes (as is the case in FEWREL).
I also perform a similar ablation for Gaussian and uniform noise, in which I generate augmented examples until each class has \naug\ $\cdot$ \nmany\ training examples. 

\cref{fig:n_aug} shows these results. 
For uniform and Gaussian noise, additional augmented examples did not further improve performance after around \naug\ $=16$.
For \methodname, on the other hand, improvement continued to increase as \naug\ increased (although the marginal improvement decreases for each \naug). 
This result confirms the intuition that extrapolations from more classes provided additional value during training.

\section{Related Work}

\vspace{-0.3em} \paragraph{Text data augmentation.}
Data augmentation methods for NLP have garnered increased interest in recent years. 
Many common techniques modify data using either token perturbations \cite{NIPS2015_5782,sennrich-etal-2016-edinburgh} or language models \cite{sennrich-etal-2016-improving,kobayashi-2018-contextual,liu-etal-2020-data,ross2021tailor}.
These techniques occur at the input-level, where all augmented data is represented by discrete tokens in natural language. 


\vspace{-0.3em} \paragraph{Hidden space augmentation.}
A growing direction in data augmentation has proposed to augment data in hidden space instead of at the input-level. 
In computer vision, \citet{devries2017dataset} explored noising, interpolation, and extrapolation, and \textsc{MixUp} \cite{mixup} combines pairs of examples. 
These methods have since been adopted to NLP---\citet{chen-etal-2020-mixtext} modify \textsc{MixUp} to improve semi-supervised text-classification, and \citet{kumar-etal-2019-closer} explore various hidden space augmentation techniques for few-shot intent classification, which I evaluated as baselines. 
Whereas the extrapolation technique used by \citet{devries2017dataset} and \citet{kumar-etal-2019-closer} (which I call ``within-extrapolation" in this paper) extrapolates the hidden space between a pair of points in the hidden space, \methodname\ extrapolates the hidden space distribution of one class onto another class.

\vspace{-0.3em} \paragraph{Example extrapolation.}
In vision, \citet{NEURIPS2018_1714726c} used a modified auto-encoder to synthesize new examples from category after seeing a few examples from it, improving few-shot object recognition.
Perhaps most similar to this work, \citet{lee2021neural} train T5 \cite{JMLR:v21:20-074} to, given some examples of a class as an input sequence, generate additional examples.
Because \methodname\ operates in hidden space, it is simpler and more computationally accessible compared with fine-tuning T5 for each classification task.

\section{Discussion}

The motivation for this work emerged from a mixture of failed experiments (I tried to devise an algorithm to select better augmented sentences in hidden space) in addition to an admiration for the elegance of the Ex2 \citep{lee2021neural}.
In hindsight, it would have been helpful to compare the performances of these two techniques in the same setting (notably, whereas I artificially restrict the sample size for certain classes in this paper, Ex2 uses the original data distributions of the datasets, which is a harder setting to show improvements from data augmentation).

I would be remiss not to mention at least one weakness that I see in my own work.
There has been an influx of recent work proposing various augmentation techniques for different NLP tasks, and due to the lack of standardized evaluation datasets and models, many papers\footnote{I'll offer up my own paper \cite{wei-zou-2019-eda} as an example. That paper should have compared with backtranslation and contextual augmentation at least.} do not perform a full comparison with respect to relevant baselines. 
This paper circumvents comparing with many data augmentation baselines (e.g., \citet{chawla2002smote}) by focusing on the question of whether hidden-space example extrapolation is a valid inductive bias (and not whether it is the best augmentation technique).
Hence, although I find example extrapolation to be a nice idea, I should concede that the particular \methodname\ operationalization of example extrapolation should undergo more comprehensive comparison with baselines before I can recommend it as a go-to augmentation technique.

In summary, I have proposed a data augmentation protocol called \methodname, which extrapolates the hidden space distribution of one class onto another.
The empirical experiments in this paper suggests that example extrapolation in hidden space is a valid inductive bias for data augmentation.
Moreover, \methodname\ is appealing because it has no hyperparameters, is model agnostic, and is lightweight.
If example extrapolation is an idea deserving of further exploration by our field, I hope this paper adds a leaf to the tree of knowledge in that space.

\section*{Acknowledgements}
Thanks Dan Garrette and Hyung Won Chung for providing feedback on the manuscript, and Barret Zoph and Ekin Dogus Cubuk for general feedback.

\bibliography{naacl2021}
\bibliographystyle{acl_natbib}

\clearpage
\newpage
\appendix

\end{document}